\documentclass{article} %
\usepackage{iclr2027_conference,times}

\usepackage{amsmath,amsfonts,bm}

\def\eqref#1{equation~\ref{#1}}

\def\1{\bm{1}}

\DeclareMathAlphabet{\mathsfit}{\encodingdefault}{\sfdefault}{m}{sl}
\SetMathAlphabet{\mathsfit}{bold}{\encodingdefault}{\sfdefault}{bx}{n}

\usepackage{hyperref}
\usepackage{url}
\usepackage{xcolor}
\usepackage{graphicx}
\usepackage{amsthm}
\usepackage{bbm}
\usepackage{multirow}
\usepackage{booktabs}
\usepackage{xspace}
\usepackage{tcolorbox}
\usepackage{enumitem}
\tcbuselibrary{most}
\newtheorem{definition}{Definition}

\usepackage{caption}
\newcommand{\xhdr}[1]{\noindent{\textbf{{#1.}}}\xspace}
\newcommand{\name}{\textsc{GuIDeR}\xspace}

\newtcolorbox{controller}{
   colback=red!5,
   colframe=red!20,
   fonttitle=\bfseries,
   title={Controller Prompt},
   arc=2mm,
   boxrule=0.3pt,
   left=1mm,
   right=1mm,
   top=1mm,
   bottom=1mm,
   enhanced,
   breakable,
   coltitle=black
}

\newtcolorbox{policy}{
   colback=red!5,
   colframe=red!20,
   fonttitle=\bfseries,
   title={Policy Prompt},
   arc=2mm,
   boxrule=0.3pt,
   left=1mm,
   right=1mm,
   top=1mm,
   bottom=1mm,
   enhanced,
   breakable,
   coltitle=black
}

\newtcolorbox{example}{
   colback=red!5,
   colframe=red!20,
   fonttitle=\bfseries,
   title={Example},
   arc=2mm,
   boxrule=0.3pt,
   left=1mm,
   right=1mm,
   top=1mm,
   bottom=1mm,
   enhanced,
   breakable,
   coltitle=black
}

\newtcolorbox{template}{
   colback=red!5,
   colframe=red!20,
   fonttitle=\bfseries,
   title={Template},
   arc=2mm,
   boxrule=0.3pt,
   left=1mm,
   right=1mm,
   top=1mm,
   bottom=1mm,
   enhanced,
   breakable,
   coltitle=black
}

\iclrfinalcopy %

\title{
Latent Decision Flow Rewards for  \\ Clinical Guideline Reasoning
}

\author{%
  Gaurav Bhatt\\
  Harvard University\\
  \texttt{gaurav\_bhatt@hms.harvard.edu} \\
  \And
  Marinka Zitnik \\
  Harvard University\\
  \texttt{marinka@hms.harvard.edu} \\
}

\begin{document}

\maketitle
\fancyhead{}
\fancyfoot{}
\cfoot{\thepage}

\begin{abstract}

Clinical practice guidelines encode evidence-based recommendations as conditional decision flows, which makes them a natural source of process supervision for training reasoning models. Two properties of clinical reasoning make this supervision difficult to use. A decision flow consistent with a guideline does not by itself determine the correct answer for a given patient, which we call the \emph{path--answer gap}, and several valid decision flows can support the same clinical decision, which we call \emph{path multiplicity}. Current training approaches address parts of this problem. Answer rewards supervise the final decision, and process rewards computed against a reference flow supervise one valid decision flow per case. Supervised fine-tuning on annotated flows teaches guideline-consistent reasoning but constrains the model to one trajectory per case and can reduce answer accuracy.
We introduce \name, a method that jointly optimizes a decision flow controller and a reasoning policy. The controller learns to generate plausible decision flows from guideline-derived supervision, and the policy learns to generate a decision flow and a final answer for each patient case. \name computes the process reward for the policy over several controller-generated flows, which accounts for path multiplicity, and gates this reward on answer correctness, which addresses the path--answer gap. Joint training allows the process supervision to evolve with the policy. The policy generates its decision flow from the patient case alone, so \name requires no guideline access at inference. On MedGUIDE, \name with Qwen3-14B achieves 82.5 weighted accuracy and 73.5 guideline adherence, 3.3 and 3.1 points above the strongest clinical reinforcement learning baseline. With Qwen3-14B and Gemma4-12B backbones, \name also improves over CRPO, MRPO, and FaithMed on the out-of-domain benchmarks AMEGA-LLM, RealMedQA, and CancerGUIDE.

\end{abstract}

\begin{center}
\name project: \url{https://zitniklab.hms.harvard.edu/GUIDER}

\name model: \url{https://github.com/mims-harvard/GUIDER}
\end{center}

\section{Introduction}

Current methods train large language models (LLMs) for clinical reasoning with rewards for final answer correctness or with preferences over complete responses \citep{jung2026mrpo}. These rewards are sparse and score each response as a single outcome. A clinical recommendation, however, depends on a sequence of conditional judgments about findings, contraindications, disease severity, and escalation thresholds, and an error at any step can change the recommendation. A model can also reach the correct answer through incomplete reasoning, with a rationale that omits essential evidence or deviates from clinical guidelines \citep{afolabi2026faithful,jiang2026better,farahani2026t2dbench,chen2026megacdp}. Clinical reasoning models should therefore learn to reach correct answers through decision pathways consistent with clinical guidelines.

\begin{figure}[t]
  \centering
  \resizebox{\linewidth}{!}{%
    \includegraphics{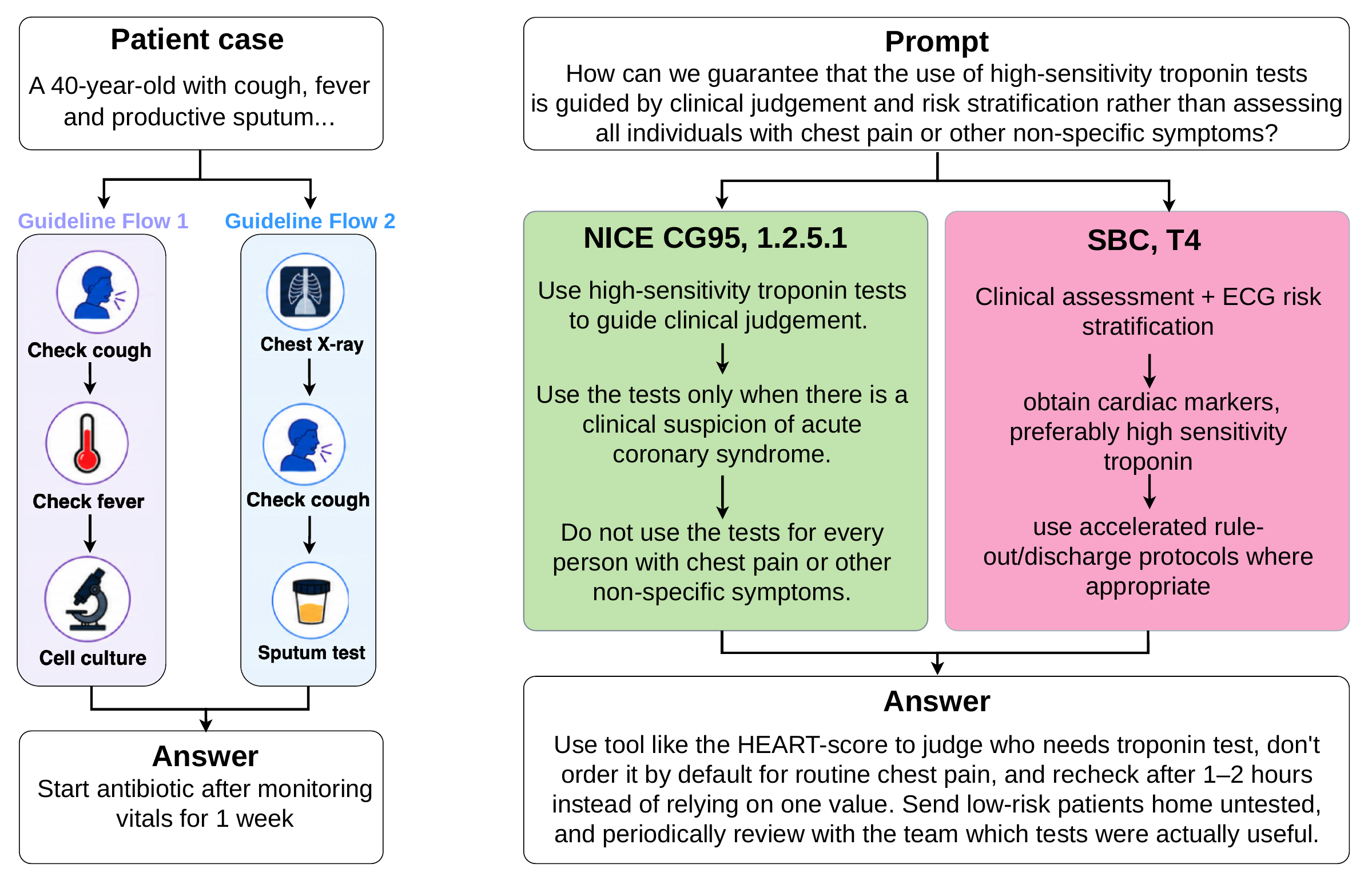}
  }
\caption{\textbf{\emph{Path multiplicity} and \emph{path--answer gap} in guideline decision flows.}
\textbf{Left.} A hypothetical patient case in which two valid decision flows lead to the same clinical answer, which illustrates path multiplicity.
\textbf{Right.} A clinical question on the use of high-sensitivity troponin testing. Decision flows from the NICE and SBC guidelines support the same answer, which illustrates path multiplicity, and the complete answer extends beyond the final step of each flow, which illustrates the path--answer gap. The NICE flow follows NICE CG95 (Recommendations~1.2.5.1--1.2.5.7) \citep{nice2016chestpain}, and the SBC flow follows the Brazilian Society of Cardiology guideline (Table~4) \citep{feitosa2015sbc}.}
  \label{fig:main}
\end{figure}

Clinical practice guidelines provide a principled source of process supervision, because they encode evidence-based recommendations as conditional decision structures \citep{li2025medguide,unell2025cancerguide,fast2024amega,kell2024realmedqa}. We call a sequence of clinical conditions and actions that leads to a clinical decision a \emph{decision flow}. Recent work compiles guidelines into executable decision logic for training \citep{shen2026medguidex} and inference \citep{cao2026guideskill}. MedGuideX \citep{shen2026medguidex}, for example, executes a deterministic decision tree for each case and trains the model on the resulting path. The decision flow behind an answer is \emph{latent}, because several decision flows are compatible with the same patient case and answer. As Figure~\ref{fig:main} shows, different guidelines can lead to the same clinical outcome through different decision flows. Two clinicians can also reach the same decision through different valid paths, based on their experience and interpretation of the evidence. We call this property \emph{path multiplicity}. Under path multiplicity, a process reward computed against a single reference flow penalizes other flows that are also consistent with guidelines. Process supervision should therefore account for the set of plausible decision flows for each case.

Reinforcement learning with verifiable rewards (RLVR) \citep{shao2024deepseekmath,deepseekai2025deepseekr1} improves clinical reasoning beyond supervised fine-tuning by optimizing verifiable outcome rewards \citep{afolabi2026faithful,jiang2026better,jung2026mrpo}. Current RLVR methods for clinical reasoning optimize answer correctness or sequence-level reasoning quality and provide limited supervision on whether intermediate clinical decisions follow established guidelines. Clinical decision flows can provide this process supervision, because they specify the relevant patient conditions, decision criteria, and recommended actions. Process supervision derived from guidelines, however, introduces a second challenge. A valid decision flow can end in a diagnostic test, an eligibility criterion, or a management action that differs from the answer that the patient-specific question requires. We call this mismatch between process validity and answer correctness the \emph{path--answer gap}. Under the path--answer gap, guideline adherence and answer correctness are distinct objectives, and a training objective should model them as complementary targets.

\xhdr{Present work}
\name (\textbf{Gu}ideline-\textbf{I}nformed \textbf{De}cision Flow Rewards for Clinical \textbf{R}easoning) is a reinforcement learning method that trains LLMs to generate guideline-consistent decision flows and correct clinical answers. To address path multiplicity, \name treats the decision flow as a latent variable and represents guideline reasoning with several plausible flows for each case. A controller trained on guideline-derived reference flows generates candidate decision flows, and \name computes the process reward as a score-weighted Monte Carlo average over these candidates. To address the path--answer gap, \name gates the process reward on answer correctness, so a policy rollout receives process credit only when its final answer is correct. \name optimizes the controller and the policy jointly. The policy generates its own decision flow from the patient case and receives guideline supervision through a scalar process reward. We train and evaluate \name on MedGUIDE \citep{li2025medguide}, an oncology benchmark derived from clinical guidelines with annotated decision paths, and evaluate transfer to the out-of-domain benchmarks CancerGUIDE \citep{unell2025cancerguide}, AMEGA-LLM \citep{fast2024amega}, and RealMedQA \citep{kell2024realmedqa}.

This paper makes three contributions.
\begin{itemize}[leftmargin=*, labelindent=0pt, topsep=0pt, itemsep=0pt]
    \item We formulate guideline decision flows as latent process supervision for clinical RLVR and identify two properties of this supervision, \emph{path multiplicity} and \emph{path--answer gap}.

    \item We introduce \name, which jointly optimizes a decision flow controller and a reasoning policy. The controller learns decision flows from guideline supervision. The controller then generates candidate decision flows, and \name computes an answer-gated process reward through score-weighted Monte Carlo aggregation.

    \item On MedGUIDE, \name achieves 82.5 weighted accuracy and 73.5 guideline adherence, 3.3 and 3.1 points above the strongest RLVR baseline. \name also achieves the highest AMEGA-LLM score among RLVR methods (22.3) and improves over clinical RLVR baselines on RealMedQA and CancerGUIDE.
\end{itemize}

\section{Related Work}
\xhdr{Clinical Guideline Benchmarks}
Clinical practice guidelines (CPGs) encode evidence-based recommendations for diagnosis and treatment as conditional decision structures, which makes them a natural basis for evaluating clinical reasoning. Recent benchmarks evaluate different aspects of reasoning grounded in guidelines. MedGUIDE derives decision-making tasks from oncology decision trees \citep{li2025medguide}, and CancerGUIDE evaluates longitudinal treatment trajectories for which several clinical pathways are plausible \citep{unell2025cancerguide}. Other benchmarks assess guideline adherence in multistage decision processes and open-ended clinical scenarios \citep{chen2026megacdp,fast2024amega}. RealMedQA \citep{kell2024realmedqa} and GuidelineQA \citep{ding2026building_silver} evaluate realistic clinical questions whose answers depend on guideline recommendations, and Meditron3 \citep{meditron3} provides a large collection of multiple-choice questions derived from the guidelines of several healthcare organizations. These benchmarks measure guideline adherence at evaluation time, and benchmarks with annotated decision paths use these paths as evaluation targets. \name uses annotated decision paths as supervision for reinforcement learning. Because several guideline pathways can be valid for one patient case, as CancerGUIDE makes explicit, \name treats decision flows as latent variables with several valid realizations per case. We train and evaluate \name on several of these benchmarks.

\xhdr{Guideline Adherence in LLMs}
A complementary line of research represents CPGs in forms that LLMs can retrieve or execute. CPGPrompt converts guidelines into decision trees that LLMs use at inference \citep{cpgprompt2026}, Guideline2Graph extracts clinical decision graphs from guideline documents \citep{guideline2graph2026}, and Case2Flow retrieves the guideline flowcharts relevant to a patient case \citep{case2flow2026}. GuideSkill compiles guideline criteria into executable skills for LLM agents \citep{cao2026guideskill}, and MEGA-CDP formulates guideline adherence as a multistage clinical decision process \citep{chen2026megacdp}. Most of these methods use retrieved, predefined, or executable CPGs at inference.
The work closest to ours is MedGuideX \citep{shen2026medguidex}, which compiles guideline recommendations into deterministic decision trees and trains LLMs on question answering data generated by executing these trees. Each case in MedGuideX has one executed path, the model imitates verbalized traces of this path during supervised fine-tuning, and the reinforcement learning stage rewards final answer correctness. Like MedGuideX, \name uses guidelines during training and operates from the patient case alone at inference. \name rewards the reasoning process directly during reinforcement learning, with a process reward aggregated over several plausible flows and gated on answer correctness, which accounts for path multiplicity and path--answer gap.

\xhdr{Verifiable and Process Rewards for Clinical Reasoning}
Reinforcement learning with verifiable rewards (RLVR) improves reasoning by optimizing against outcomes that can be checked automatically \citep{shao2024deepseekmath,deepseekai2025deepseekr1}, and process reward models (PRMs) provide denser supervision over intermediate reasoning steps \citep{uesato2022processoutcome,lightman2024letsverify,mathshepherd2024}. Recent clinical methods use several forms of process reward. CRPO jointly rewards accuracy, faithfulness, and comprehensiveness \citep{gu2025clinicalr1}, MRPO introduces step-aware rewards that penalize invalid intermediate reasoning \citep{jung2026mrpo}, and FaithMed scores reasoning and retrieval steps against clinician-designed criteria \citep{zhang2026faithmed}. These methods score reasoning traces against predefined criteria, and \name derives its process reward from the conditional structure of clinical guidelines.
Process rewards derived from guidelines must account for two properties of clinical reasoning. Under path multiplicity, several guideline-consistent flows can support the same outcome, and a reward computed against a single reference flow penalizes valid alternative flows. Under path--answer gap, a valid flow can end in a step that differs from the answer for a given patient, so process and outcome rewards can disagree. \name addresses path multiplicity by aggregating process rewards over several candidate flows from a controller trained on guideline-derived supervision. It addresses path--answer gap by gating this reward on answer correctness, so that only rollouts with a correct final answer receive process credit.

\section{Method}
\subsection{Problem Formulation}
\label{sec:prob_formulation}

\xhdr{Setup and Notation}
Let $x \in \mathcal{X}$ denote a patient case and $y^\ast \in \mathcal{Y}$ its target clinical answer. Each training pair $(x,y^\ast)$ is associated with a clinical practice guideline $G$, from which we derive a reference decision flow $z^\ast$ for training supervision. The policy $\pi_\theta$ with parameters $\theta$ generates a decision flow $\hat z$ and a final answer $\hat y$ from the patient case, $(\hat z,\hat y)\sim\pi_\theta(\cdot\mid x)$. A decision flow controller $q_\theta$ generates candidate decision flows and provides process supervision to the policy during training. The controller and the policy share parameters $\theta$, differ in their prompts, and are optimized jointly. We write $z^\ast$ for the annotated reference flow, $z$ for a controller-generated flow, and $\hat z$ for a policy-generated flow. At inference, \name uses only policy $\pi_\theta(\cdot\mid x)$, which generates the decision flow and the answer from the patient case alone. The method therefore targets guideline-aligned decision making, and guideline retrieval and identification are outside its scope.

\begin{definition}[Clinical Decision Flow]
Given a patient case $x$ and a clinical guideline $G$, a \emph{clinical decision flow} is a sequence
\begin{equation}
    z = (v_1, v_2, \ldots, v_T),
\end{equation}
where each $v_t$ is a clinical condition, criterion, decision, or action derived from $G$. A transition $(v_t,v_{t+1})$ is valid if it is consistent with the guideline logic in $G$ and with the patient state described by $x$. We denote the set of valid decision flows for $x$ under $G$ by $\mathcal{Z}(x;G)$.
\end{definition}

The learning objective requires a clinically correct final answer $\hat y$ and a decision flow $\hat z$ consistent with the relevant guideline.

\xhdr{Path Multiplicity}
Clinical guideline reasoning is a many-to-one problem because several valid decision flows can support the same clinical answer. For a patient--outcome pair $(x,y^\ast)$, there can exist flows
\begin{equation}
    z_i \neq z_j, \qquad
    z_i,z_j \in \mathcal{Z}(x;G),
    \qquad
    y^\ast \in
    \mathcal{Y}(x,z_i) \cap \mathcal{Y}(x,z_j),
\end{equation}
where $\mathcal{Y}(x,z)$ denotes the set of clinical answers compatible with patient $x$ and decision flow $z$. Supervision against a single annotated trajectory can therefore penalize alternative flows that are clinically valid.

\begin{definition}[Latent Decision Flow]
The decision flow underlying an answer is \emph{latent} when it is unobserved in the patient--outcome pair $(x,y^\ast)$ and the set $\mathcal{Z}(x;G)$ contains several guideline-consistent flows that support $y^\ast$.
\end{definition}

\xhdr{Path--Answer Gap}
A guideline-consistent decision flow can end in a step that differs from the answer the task requires. Its terminal node $v_T$ can be a clinical state, diagnostic test, eligibility criterion, or management action, so in general $v_T \not\equiv y^\ast$. Answer correctness ($\hat y = y^\ast$) and flow validity ($\hat z \in \mathcal{Z}(x;G)$) are therefore distinct objectives. A policy can produce the correct answer through a flow outside $\mathcal{Z}(x;G)$, or generate a valid flow that leads to an incorrect answer.

\xhdr{Clinical Intuition}
Consider the chest pain example in Figure~\ref{fig:main}, in which the clinical question asks when clinicians should use high-sensitivity troponin testing. A decision flow for this question represents the guideline-derived clinical considerations that lead to the recommendation, such as the assessment of clinical suspicion for acute coronary syndrome, risk stratification, and the indication for troponin testing. The NICE and SBC guidelines express distinct but clinically consistent sequences that support the same recommendation, which illustrates \emph{path multiplicity}. Because the patient case and the final answer are compatible with several valid flows, the decision flow is \emph{latent}. The complete answer extends beyond the final step of each guideline flow and follows from the clinical decisions along the flow, which illustrates the \emph{path--answer gap}. These properties motivate a training objective that accounts for several plausible decision flows and jointly optimizes flow validity and answer correctness.

\subsection{\name Model}

\begin{figure}[t]
    \centering
    \includegraphics[width=\linewidth]{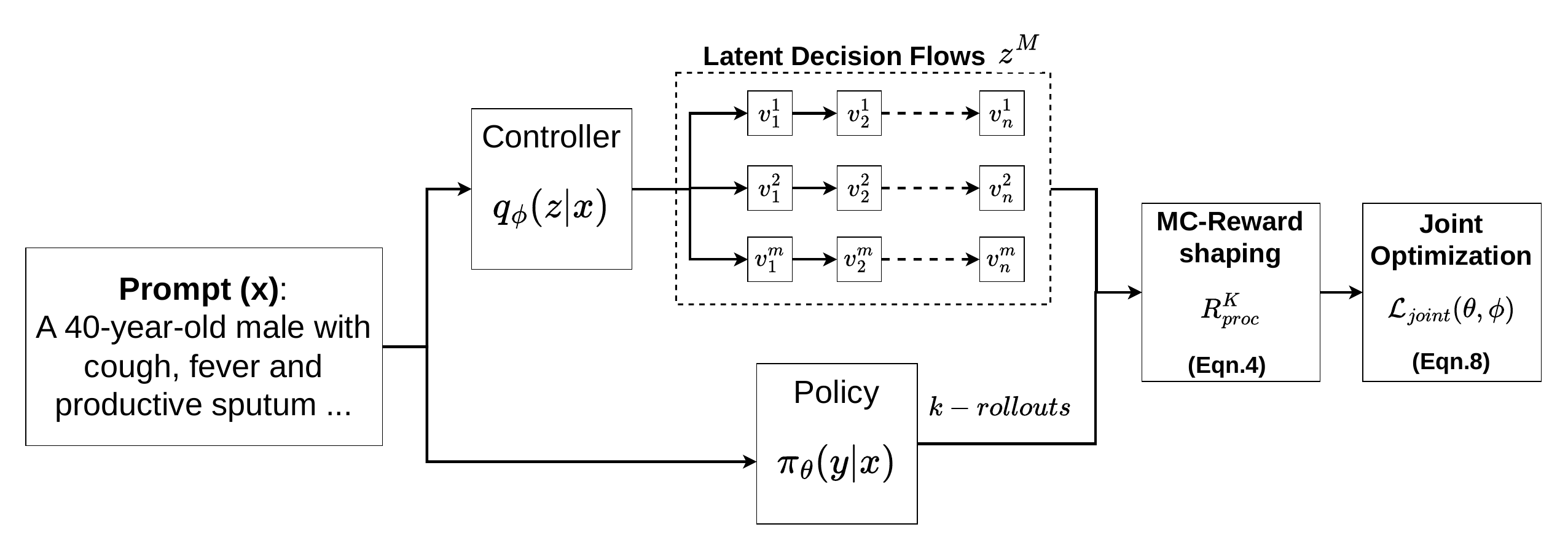}
    \caption{\textbf{Overview of \name.}
    The controller learns to generate decision flows from guideline-derived reference flows and samples several candidate flows for each patient case. The policy generates its own decision flow and final answer from the patient case. \name computes an answer-gated process reward from the agreement between the policy flow and the controller flows, and it optimizes the policy with GRPO and the controller with guideline supervision in a joint objective.}
    \label{fig:method}
\end{figure}

Figure~\ref{fig:method} gives an overview of \name, which trains a controller and a policy jointly. Given a patient case $x$, the controller generates several candidate decision flows, and the policy generates its own decision flow and final answer. \name aggregates the agreement between the policy flow and the controller flows into a process reward, which accounts for several plausible paths for each case. \name combines this process reward with answer correctness to optimize the policy and trains the controller on guideline-derived reference flows. Through this joint optimization, \name transfers guideline supervision to the policy, and the policy generates decision flows from the patient case alone at inference.

We describe \name in three parts. The \emph{decision flow controller} learns to generate candidate decision flows from guideline-derived supervision, the \emph{latent decision flow reward} aggregates process credit across these candidates, and \emph{joint controller and policy optimization} trains the model for guideline-consistent reasoning and answer correctness.

\subsubsection{Decision Flow Controller}

The decision flow controller learns to generate candidate decision flows from guideline-derived supervision. During training, we pair each case with an annotated reference flow $z^\ast=(v_1^\ast,\ldots,v_T^\ast)$ derived from its guideline $G$. We train the controller with the autoregressive negative log-likelihood $
    \mathcal{L}_{\mathrm{ctrl}}(\theta)
    =
    -\sum_{t=1}^{T}
    \log q_\theta
    \left(
        v_t^\ast \mid x,v_{<t}^\ast
    \right),
    \label{eq:controller}$
where $v_{<t}^\ast=(v_1^\ast,\ldots,v_{t-1}^\ast)$ denotes the reference flow prefix. The reference flow $z^\ast$ supervises the controller, and the policy receives guideline supervision through the process reward. Sampling from the controller yields several candidate flows for each case, which allows the process reward to account for path multiplicity.

\subsubsection{Latent Decision Flow Reward}

\xhdr{Process Supervision over Multiple Flows}
Under path multiplicity, supervision against a single annotated flow penalizes valid alternative flows. \name therefore compares the policy flow with several candidate flows generated by the controller. At training step $t$, we sample $
    z^{(m)} \sim q_{\theta_t}(z\mid x),\
     m=1,\ldots,M,
    \label{eq:controller_sampling}$ where $\theta_t$ denotes the shared model parameters before the update at step $t$.

We score each controller proposal by its coverage of the annotated reference flow. Let $\mathcal{V}(z^\ast)$ denote the set of reference decision nodes, and let $\operatorname{match}(v,z)$ indicate whether flow $z$ matches reference node $v$. The coverage score of proposal $m$ is
\begin{equation}
    s_m
    =
    \lambda_{\mathrm{PRM}}
    \frac{1}{|\mathcal{V}(z^\ast)|}
    \sum_{v\in\mathcal{V}(z^\ast)}
    \operatorname{match}(v,z^{(m)}),
    \label{eq:controller_path_score}
\end{equation}
where $\lambda_{\mathrm{PRM}}$ scales the coverage score. A softmax over the scores gives the normalized proposal weights
$w_m
    =
    \frac{\exp(s_m/\tau)}
    {\sum_{j=1}^{M}\exp(s_j/\tau)},
    \ m=1,\ldots,M,
    \label{eq:mc_weights}
$ where the temperature $\tau>0$ controls how strongly the weights concentrate on proposals with higher coverage.

\xhdr{Monte Carlo Process Reward}
We measure agreement between the policy flow and a controller proposal with token-set Jaccard similarity. Let $\operatorname{Tok}(z)$ denote the set of alphanumeric tokens in $z$ after lowercasing, stopword removal, and removal of tokens with at most two characters. The similarity is
$
    J(z_1,z_2)
    =
    \frac{
        |\operatorname{Tok}(z_1)\cap\operatorname{Tok}(z_2)|
    }{
        |\operatorname{Tok}(z_1)\cup\operatorname{Tok}(z_2)|
    },
    \label{eq:flow_jaccard}
$
with $J(z_1,z_2)=0$ when either token set is empty.

For a policy-generated flow $\hat z$, each controller proposal contributes process credit in proportion to its reference coverage and its agreement with $\hat z$. The process reward is
\begin{equation}
    R_{\mathrm{proc}}(\hat z)
    =
    \sum_{m=1}^{M}
    w_m\,[s_m]_+\,
    J\!\left(\hat z,z^{(m)}\right),
    \qquad
    [a]_+=\max(0,a).
    \label{eq:process_reward}
\end{equation}
This reward is the expected proposal-specific process credit under the categorical distribution defined by the weights $w_m$ (Appendix~\ref{app:multiflow_reward}). Controller proposals with higher reference coverage contribute more to the reward, and the policy receives process credit in proportion to its agreement with this weighted set of candidate flows.

\xhdr{Answer-Gated Reward Shaping}
Multi-flow process supervision accounts for \emph{path multiplicity}, and the \emph{path--answer gap} requires a second mechanism, because a guideline-consistent decision flow can lead to an incorrect clinical answer. We therefore combine the process reward with the verifiable answer reward $R_{\mathrm{ans}}(\hat y,y^\ast)=\mathbbm{1}[\hat y=y^\ast]$. The reward for a policy-generated completion $(\hat z,\hat y)$ is
$
    R(\hat z,\hat y)
    =
    R_{\mathrm{ans}}(\hat y,y^\ast)
    +
    \lambda
    R_{\mathrm{ans}}(\hat y,y^\ast)
    R_{\mathrm{proc}}(\hat z),
    \label{eq:shaped_reward}
$
where $\lambda\geq0$ controls the contribution of process supervision. The answer gate gives process credit only to completions with a correct clinical answer. Multi-flow process supervision accounts for several plausible decision flows, and answer-gated reward shaping couples guideline-consistent reasoning with answer correctness, which addresses the two properties identified in Section~\ref{sec:prob_formulation}.

\subsubsection{Joint Controller and Policy Optimization}

\name combines controller supervision and policy optimization in a single objective over shared parameters. Guideline-derived reference flows train the controller, and answer-gated process rewards aggregated over several controller-generated flows train the policy. The two roles share the same model parameters and receive joint updates.

At training step $t$, the shared model has parameters $\theta_t$. Before any parameter update at step $t$, this model generates $M$ controller flows and $K$ policy rollouts,
\begin{equation}
    z^{(m)} \sim q_{\theta_t}(z\mid x),
    \qquad
    (\hat z^{(k)},\hat y^{(k)})
    \sim \pi_{\theta_t}(\cdot\mid x),
    \label{eq:joint_sampling}
\end{equation}
for $m=1,\ldots,M$ and $k=1,\ldots,K$.

For each controller proposal $z^{(m)}$, we compute its coverage score $s_m$ with Eq.~\ref{eq:controller_path_score}. The process reward for policy rollout $k$ is
\begin{equation}
    R_{\mathrm{proc}}^{(k)}
    =
    \sum_{m=1}^{M}
    w_m [s_m]_+
    J\!\left(\hat z^{(k)},z^{(m)}\right),
    \label{eq:rollout_process_reward}
\end{equation}
where $J(\cdot,\cdot)$ measures agreement between the policy flow and each controller proposal, and the answer-gated reward is
\begin{equation}
    R^{(k)}
    =
    R_{\mathrm{ans}}^{(k)}
    +
    \lambda
    R_{\mathrm{ans}}^{(k)}
    R_{\mathrm{proc}}^{(k)},
    \qquad
    R_{\mathrm{ans}}^{(k)}
    =
    \mathbbm{1}[\hat y^{(k)}=y^\ast].
    \label{eq:rollout_reward}
\end{equation}

Following GRPO \citep{shao2024deepseekmath}, we normalize the rewards within the group of $K$ rollouts,
$
    A^{(k)}
    =
    \frac{R^{(k)}-\mu_R}{\sigma_R+\epsilon},
    \ k=1,\ldots,K,
    \label{eq:grpo_advantage}
$ where $\mu_R$ and $\sigma_R$ are the mean and standard deviation of the $K$ rollout rewards, and use these advantages in the standard GRPO objective $\mathcal{L}_{\mathrm{GRPO}}$.

Because the controller and the policy share parameters, we combine their objectives before a single optimizer update,
\begin{equation}
    \mathcal{L}_{\mathrm{joint}}(\theta_t)
    =
    \mathcal{L}_{\mathrm{GRPO}}
    \left(
        \theta_t;\{R^{(k)}\}_{k=1}^{K}
    \right)
    +
    \alpha_t
    \mathcal{L}_{\mathrm{ctrl}}(\theta_t),
    \label{eq:joint}
\end{equation}
where $\alpha_t$ controls the contribution of controller supervision. A single optimizer step on $\mathcal{L}_{\mathrm{joint}}$ produces $\theta_{t+1}$. Controller proposals, policy rollouts, and rewards stay fixed during this update, so the two objectives are computed from the same pre-update model $\theta_t$.

Guideline-derived flow supervision and policy optimization thus train a single model, and the multi-flow process reward credits several plausible flows for each case. At inference, \name uses only the policy $\pi_\theta(\cdot\mid x)$, which generates the decision flow and the answer from the patient case.

\section{Experiments}
\subsection{Experimental Setup}

\xhdr{Datasets and Performance Metrics}
We train on the MedGUIDE training split \citep{li2025medguide} and a subset of
Meditron3 (GuidelineQA) \citep{meditron3}. We evaluate on held-out MedGUIDE and
cross-benchmark on AMEGA-LLM \citep{fast2024amega}, RealMedQA
\citep{kell2024realmedqa}, and CancerGUIDE \citep{unell2025cancerguide}.

We follow the evaluation protocol and judge templates provided by each
benchmark, using GPT-5.4 as the LLM judge where required. For MedGUIDE, we
report weighted accuracy (WAcc.) following \citet{li2025medguide} and
guideline adherence on a 0--5 scale. For AMEGA-LLM, we report the
criterion-based aggregate score (0--50). For RealMedQA, we report the mean
flow-judge score (0--5), and for CancerGUIDE, the mean clinical-agreement
score (0--5). The benchmark-provided evaluation templates and scoring rubrics
are reproduced in Appendix~\ref{app:evaluation_prompts}.

\xhdr{Baselines}
We compare against medical foundation models, recent clinical reinforcement learning methods, and frontier LLMs. The medical models are Meditron-3 \citep{meditron3} and Fastino-Healthcare \citep{fastnemo}. Our primary baselines are Clinical-R1 (CRPO) \citep{gu2025clinicalr1}, which uses multi-objective verifiable rewards, MRPO \citep{jung2026mrpo}, which uses step-wise process rewards, and FaithMed \citep{zhang2026faithmed}, which uses rubric-based process supervision for reasoning grounded in evidence. We also report GPT-5.6-Sol and Gemini-Flash-3.7 as frontier references.

\xhdr{Implementation Details}
We use Qwen3-14B as our primary backbone and Gemma4-12B as a second backbone.
Models are fine-tuned with QLoRA using 4-bit quantization, bfloat16
computation, LoRA rank $64$, scaling factor $128$, and dropout $0.05$.
Answer-only SFT is trained for one epoch with learning rate $2\times10^{-4}$;
GRPO and joint optimization are initialized from this checkpoint. For GRPO and joint optimization, we sample $K=4$ completions, use
$\beta=0.04$, learning rate $5\times10^{-6}$, batch size $16$, maximum prompt
length $2048$, and disable thinking. Joint-MC policy completions are limited
to $256$ tokens. For joint optimization, we set $\alpha_t=1.0$ and
$\lambda=0.2$, and sample $M=4$ controller flows with temperature $0.7$ and
top-$p=0.9$, limited to four decision nodes and $128$ tokens. We set the
reweighting temperature to $\tau=0.2$ and use the score-weighted MC process
reward in Eq.~\ref{eq:process_reward}. Joint-MC is trained for one epoch (approximately 655 updates on 10,466 examples) on a single NVIDIA H200 GPU. Evaluation uses deterministic decoding with thinking disabled.  Unless otherwise stated, we report the mean and standard deviation over three independent runs with different random seeds. All methods use the same training data and initialization when applicable.

\subsection{Results}

\begin{table*}[t]
  \centering
\caption{\textbf{Comparison of training strategies with Qwen3-14B.}
We compare the base model, supervised fine-tuning (SFT) and GRPO with answer supervision, SFT and GRPO with answer and decision flow supervision, and \name. ``ans'' denotes supervision on the final answer, which is the answer target for SFT and the answer correctness reward for GRPO. ``ans+flow'' adds supervision on the annotated reference decision flow, which is the flow target for SFT and a process reward computed against the reference flow for GRPO. The arrow denotes GRPO initialized from the corresponding SFT checkpoint. We report weighted accuracy (WAcc.) and guideline adherence (Adh.) on MedGUIDE, the aggregate score on AMEGA-LLM, and LLM judge scores on RealMedQA and CancerGUIDE. Higher is better for all metrics.}
  \label{tab:ablation}

  \resizebox{\textwidth}{!}{%
  \begin{tabular}{lccccc}
  \toprule
  \multirow{2}{*}{Training Strategy}
  & \multicolumn{2}{c}{MedGUIDE}
  & \multirow{2}{*}{AMEGA-LLM}
  & \multirow{2}{*}{RealMedQA}
  & \multirow{2}{*}{CancerGUIDE} \\
  \cmidrule(lr){2-3}
  & WAcc. & Adh.
  & & & \\
  \midrule

  Base (Qwen3-14B)
  & 52.3 & 47.2 & 17.2 & 1.62 & 1.71 \\

  SFT (ans)
  & 78.0 & 53.1 & 20.3 & 1.73 & 1.73 \\

  SFT (ans) $\rightarrow$ GRPO (ans)
  & 80.1 & 54.5 & 20.5 & 1.81 & 1.75 \\

  SFT (ans+flow)
  & 64.0 & 71.3 & 18.8 & 1.82 & 1.80 \\

  SFT (ans+flow) $\rightarrow$ GRPO (ans+flow)
  & 68.2 & 71.6 & 19.1 & 1.85 & 1.82 \\

  \name
  & \textbf{82.5} & \textbf{73.5} & \textbf{22.3} & \textbf{1.90} & \textbf{2.40} \\

  \bottomrule
  \end{tabular}%
  }
\end{table*}

Table~\ref{tab:ablation} answers three questions about incorporating guideline decision flows into training.

\textbf{Q1: Can process rewards alone shape RL toward guideline adherence?}
Only to a limited extent. As shown in Table~\ref{tab:ablation}, process-reward optimization yields only modest improvements in guideline adherence (SFT (ans+flow) $\rightarrow$ GRPO (ans+flow): 71.3$\rightarrow$71.6), while answer performance remains substantially below answer-oriented training (68.2 vs.\ 80.1 WAcc.). This limitation arises because process rewards can only assign useful credit to decision flows that the policy is able to explore; without sufficient in-domain guideline knowledge, the base policy has limited access to high-quality guideline-consistent trajectories. The problem is further compounded by \emph{path multiplicity}: multiple decision flows may be valid for the same clinical outcome, making reward assignment against a single reference flow overly restrictive. Thus, process rewards alone are insufficient for reliably acquiring guideline reasoning from scratch, motivating the use of supervised guideline knowledge to initialize the policy.

\textbf{Q2: Can we first learn guideline decision flows with SFT and then optimize the resulting policy with RL?}
A natural solution to the exploration limitation above is to first inject guideline knowledge through supervised fine-tuning. Indeed, flow+answer SFT substantially improves guideline adherence over answer-only SFT (53.1$\rightarrow$71.3), demonstrating that the policy can acquire guideline-consistent decision patterns from supervised trajectories. However, this gain comes at a substantial cost in weighted answer accuracy (78.0$\rightarrow$64.0). Subsequent GRPO preserves the adherence gain (71.6) but only partially recovers answer accuracy (68.2). This divergence exposes the \emph{path--answer gap}: learning to follow a guideline-consistent decision flow does not necessarily improve the final clinical answer. Moreover, committing the policy to a single supervised trajectory remains problematic under path multiplicity. Thus, sequentially learning flows through SFT and subsequently optimizing with RL does not fully reconcile process quality with outcome correctness.

\textbf{Q3: Can we jointly optimize guideline reasoning and answer correctness?}
The limitations above motivate learning the decision process and clinical outcome jointly rather than in separate stages. Our joint MC formulation couples policy optimization with a controller that generates multiple candidate decision flows, allowing process supervision to
evolve alongside the policy while accounting for path multiplicity through
score-weighted aggregation. This avoids requiring the initial policy to discover high-quality guideline flows on its own and avoids treating a single annotated trajectory as the unique reasoning target. Joint optimization consequently achieves the strongest combination of MedGUIDE weighted accuracy (82.5) and guideline adherence (73.5), while also providing the strongest overall transfer across the evaluated out-of-domain benchmarks. These results support jointly optimizing process and outcome signals while explicitly accounting for the one-to-many relationship between clinical answers and valid guideline decision flows.

\begin{table*}[t]
  \centering
  \caption{\textbf{Performance across clinical reasoning benchmarks}. For
  fine-tuned methods, we report mean $\pm$ standard deviation over three
  random seeds. We report weighted accuracy (WAcc.) and guideline adherence
  (Adh.) on MedGUIDE, aggregate score on AMEGA-LLM, and LLM-judge scores on
  RealMedQA and CancerGUIDE. Higher is better for all metrics.}
  \label{tab:main_results_1}

  \resizebox{\textwidth}{!}{%
  \begin{tabular}{lccccc}
  \toprule
  \multirow{2}{*}{Model}
  & \multicolumn{2}{c}{MedGUIDE}
  & \multirow{2}{*}{AMEGA-LLM}
  & \multirow{2}{*}{RealMedQA}
  & \multirow{2}{*}{CancerGUIDE} \\
  \cmidrule(lr){2-3}
  & WAcc. & Adh. & & & \\
  \midrule

  GPT-5.6-Sol
  & 70.0 & 67.2 & 19.1 & \textbf{2.8} & 2.6 \\
  Gemini-Flash-3.7
  & 67.2 & 61.5 & 18.8 & 2.4 & 2.3 \\

  \midrule
  Meditron-3-8B \citep{meditron3}
  & 34.7 & 62.3 & 18.3 & 1.5 & 1.9 \\
  Fastino-Healthcare-30B \citep{fastnemo}
  & 36.7 & 41.3 & 16.1 & 1.6 & 1.3 \\

  \midrule
  \multicolumn{6}{c}{\textit{\textbf{Gemma4-12B}}} \\
  CRPO \citep{gu2025clinicalr1}
  & $62.1{\pm}0.6$ & $63.5{\pm}0.7$ & $18.2{\pm}0.3$ & $1.2{\pm}0.1$ & $1.9{\pm}0.1$ \\
  FaithMed \citep{zhang2026faithmed}
  & $63.5{\pm}0.5$ & $55.3{\pm}0.8$ & $19.4{\pm}0.2$ & $1.3{\pm}0.1$ & $2.1{\pm}0.1$ \\
  MRPO \citep{jung2026mrpo}
  & $63.9{\pm}0.6$ & $55.6{\pm}0.7$ & $19.3{\pm}0.3$ & $1.4{\pm}0.1$ & $2.4{\pm}0.1$ \\
  \name
  & $\mathbf{67.5{\pm}0.5}$ & $\mathbf{65.2{\pm}0.6}$ & $\mathbf{20.2{\pm}0.2}$ & $\mathbf{1.7{\pm}0.1}$ & $\mathbf{2.8{\pm}0.1}$ \\

  \midrule
  \multicolumn{6}{c}{\textit{\textbf{Qwen3-14B}}} \\
  CRPO \citep{gu2025clinicalr1}
  & $75.3{\pm}0.5$ & $70.4{\pm}0.7$ & $21.2{\pm}0.2$ & $1.6{\pm}0.1$ & $1.9{\pm}0.1$ \\
  FaithMed \citep{zhang2026faithmed}
  & $77.8{\pm}0.4$ & $66.9{\pm}0.6$ & $20.7{\pm}0.3$ & $1.7{\pm}0.1$ & $2.1{\pm}0.1$ \\
  MRPO \citep{jung2026mrpo}
  & $79.2{\pm}0.5$ & $67.5{\pm}0.6$ & $21.9{\pm}0.2$ & $1.8{\pm}0.1$ & $2.3{\pm}0.1$ \\
  \name
  & $\mathbf{82.5{\pm}0.4}$ & $\mathbf{73.5{\pm}0.5}$ & $\mathbf{22.3{\pm}0.2}$ & $\mathbf{1.9{\pm}0.1}$ & $\mathbf{2.4{\pm}0.1}$ \\

  \bottomrule
  \end{tabular}%
  }
\end{table*}

\xhdr{Comparison with baselines}
Table~\ref{tab:main_results_1} compares \name against frontier and open-source medical LLMs, as well as recent clinical reasoning optimization methods. \name consistently outperforms CRPO, FaithMed, and MRPO across both model families. On Qwen3-14B, \name achieves the highest MedGUIDE weighted accuracy and guideline adherence (82.5/73.5), improving over the strongest baseline by 3.3 and 3.1 points, respectively. These gains extend beyond the in-domain benchmark: \name also obtains the highest AMEGA-LLM score (22.3) and improves over all three baselines on RealMedQA and CancerGUIDE. A similar pattern holds for Gemma4-12B, where \name improves both answer accuracy and guideline adherence while providing stronger overall out-of-domain performance. Notably, the gains are not explained by using a medically specialized backbone: \name substantially outperforms Meditron-3-8B and Fastino-Healthcare-30B despite operating on general-purpose base models. Finally, on Qwen3-14B, \name surpasses GPT-5.6-Sol on MedGUIDE weighted accuracy (82.5 vs.\ 70.0) and adherence (73.5 vs.\ 67.2), while GPT-5.6-Sol remains stronger on RealMedQA (2.8 vs.\ 1.9). Overall, these results show that the improvements are consistent across model scales and extend from guideline adherence to answer accuracy and out-of-domain clinical reasoning.

\section{Discussion}

We introduce \name, a reinforcement learning method that uses clinical practice guidelines as process supervision for reasoning models. In our experiments, guideline adherence and answer correctness behave as distinct training objectives. Supervised training on annotated decision flows improves adherence and reduces answer accuracy, and answer rewards improve accuracy with smaller gains in adherence. By computing the process reward over several plausible decision flows and gating it on answer correctness, \name improves the two objectives together, and these improvements extend to out-of-domain clinical benchmarks.

\xhdr{Limitations and Future Work}
\name shares parameters between the controller and the policy. We chose this design because the base policy has limited in-domain guideline knowledge, and training the shared model on reference flows provides this knowledge during reinforcement learning. The quality of controller flows is therefore bounded by the guideline knowledge of the shared model, and errors in these flows enter the process reward. A separate controller with stronger guideline-specific knowledge could provide more accurate candidate flows and a more reliable process reward. Sampling $M$ controller flows for each case also adds computational cost to each training step, and this cost grows with $M$ and with flow length. More efficient proposal generation and selection, for example by adapting $M$ to the uncertainty of each case, could reduce this cost. The process reward measures agreement between flows with token-set Jaccard similarity, which captures lexical overlap and can miss clinically equivalent decision states expressed in different terms. Semantic matching or matching validated by clinicians could give a more accurate measure of agreement. Finally, training data come mainly from oncology guidelines, several evaluation metrics rely on LLM judges, and clinical safety requires evaluation by clinicians. Extending \name to other specialties and clinical settings, and evaluating it with clinicians, are directions for future work.

\xhdr{Conclusion}
\name jointly optimizes a decision flow controller and a reasoning policy and uses clinical decision flows as process supervision for reinforcement learning. Across in-domain and out-of-domain benchmarks and two model families, \name improves answer accuracy and guideline adherence over answer-only training, supervised flow training, and clinical process reward baselines. These results indicate that structured clinical guidelines can supervise the training of clinical reasoning models, and the trained policy generates guideline-consistent decision flows from the patient case alone at inference. More broadly, a model that states its decision flow in guideline terms allows clinicians to check each step of a recommendation against the relevant guideline, and the same training approach applies to other domains that encode expert knowledge as conditional decision procedures with several valid paths to the same decision.

\subsection*{AI Use Statement}
In this work, we used generative AI tools to assist with language editing, improving the clarity and presentation of the manuscript, and supporting code development and debugging. Generative AI was not used to autonomously conduct experiments, determine research conclusions, or evaluate the proposed method. All AI-assisted text and code were reviewed by the authors, and experimental results, analyses, citations, and scientific claims were independently verified. The authors take full responsibility for the final content of this work, including any text, claims, or artifacts produced with the aid of generative AI.

\subsection*{Ethics Statement}

\name is a research method, and its use in clinical care would require clinical validation. This work uses benchmark datasets and involves no human subjects or identifiable patient information. The main risk is that a model trained to follow guidelines can produce fluent but incorrect or outdated recommendations. Our evaluation metrics, several of which rely on LLM judges, measure agreement with reference answers and guideline paths, and clinical safety requires separate assessment by clinicians. The guidelines used in this work come mainly from high-income countries and can encode contested demographic factors. Fairness across patient subgroups is outside the scope of our evaluation and requires further study. Clinical implementation would require prospective validation and oversight by clinicians.

\subsection*{Acknowledgments}
We gratefully acknowledge support, in part, by NSF CAREER 2339524, Harvard Data Science Initiative, Amazon Faculty Research, Google Research Scholar Program, AstraZeneca Research, Roche Alliance with Distinguished Scientists (ROADS) Program, Sanofi iDEA-iTECH Award, GlaxoSmithKline Award, Boehringer Ingelheim Award, Merck Award, Optum AI Research Collaboration Award, Pfizer Research, Gates Foundation (INV-079038), Chan Zuckerberg Initiative, John and Virginia Kaneb Fellowship at Harvard Medical School, Biswas Computational Biology Initiative in partnership with the Milken Institute, Harvard Medical School Dean’s Innovation Fund for the Use of Artificial Intelligence, and the Kempner Institute for the Study of Natural and Artificial Intelligence at Harvard University. This work was delivered as part of the Aurora project supported by the Cancer Grand Challenges partnership funded by Cancer Research UK (CGCAI1-Mar26/100001).
Any opinions, findings, conclusions or recommendations expressed in this material are those of the authors and do not necessarily reflect the views of the funders.

\bibliography{iclr2027_conference}
\bibliographystyle{iclr2027_conference}

\clearpage

\appendix
\section{Appendix}

\subsection{Derivation of the Multi-Flow Process Reward}
\label{app:multiflow_reward}

At training step $t$, we draw $M$ candidate decision flows from the
controller role of the shared model,
\begin{equation}
    z^{(m)} \sim q_{\theta_t}(z\mid x),
    \qquad m=1,\ldots,M.
\end{equation}
Each proposal receives a reference-path coverage score $s_m$ as defined in
\eqref{eq:controller_path_score}. We use these scores to define a categorical
distribution over the sampled proposals,
\begin{equation}
    w_m
    =
    \frac{\exp(s_m/\tau)}
    {\sum_{j=1}^{M}\exp(s_j/\tau)},
    \qquad
    \sum_{m=1}^{M} w_m = 1.
    \label{eq:app_mc_weights}
\end{equation}
Hence, proposals with greater coverage of the annotated reference flow
receive greater mass, while lower-scoring proposals can still contribute to
the reward.

For a policy-generated flow $\hat z$, define the proposal-specific process
credit
\begin{equation}
    r_m(\hat z)
    =
    [s_m]_+\,
    J\!\left(\hat z,z^{(m)}\right),
    \label{eq:app_candidate_reward}
\end{equation}
where $J$ is the token-set Jaccard similarity defined in
\eqref{eq:process_reward}. The process reward is the expectation of this
credit under the score-induced categorical distribution over the sampled
controller proposals:
\begin{align}
    R_{\mathrm{proc}}(\hat z)
    &=
    \mathbb{E}_{m\sim\operatorname{Cat}(w_1,\ldots,w_M)}
    \left[r_m(\hat z)\right]
    \\
    &=
    \sum_{m=1}^{M}
    w_m [s_m]_+
    J\!\left(\hat z,z^{(m)}\right).
    \label{eq:app_mc_process_reward}
\end{align}

Substituting \eqref{eq:app_mc_weights} gives the equivalent form
\begin{equation}
    R_{\mathrm{proc}}(\hat z)
    =
    \frac{
        \sum_{m=1}^{M}
        \exp(s_m/\tau)\,[s_m]_+\,
        J\!\left(\hat z,z^{(m)}\right)
    }{
        \sum_{j=1}^{M}\exp(s_j/\tau)
    }.
    \label{eq:app_mc_process_expanded}
\end{equation}

This construction separates two sources of process information:
$s_m$ measures the reference-path support of each controller proposal, while
$J(\hat z,z^{(m)})$ measures its agreement with the policy-generated flow.
Together, they provide score-weighted process supervision over multiple
controller-generated flow candidates rather than requiring exact agreement
with a single annotated reference flow.

\subsection{Effect of the Number of Decision Flows}
We study the effect of the number of controller-sampled decision flows $M$ used to estimate the process reward. Increasing $M$ from 2 to 4 improves MedGUIDE weighted accuracy from 80.3 to 82.5 and adherence from 72.5 to 73.5, with similar gains across the out-of-domain benchmarks. Beyond $M=4$, performance largely saturates: weighted accuracy remains stable around 82.3--82.4, while adherence continues to improve modestly to 74.5 at $M=8$. These results suggest that a small set of sampled flows is sufficient to capture path multiplicity, with additional samples providing diminishing returns.

\subsection{Effect of Decision-Flow Length}
We study the effect of the maximum number of decision nodes $N$ allowed in each controller-sampled flow. Increasing $N$ from 2 to 4 substantially improves both MedGUIDE weighted accuracy (77.2 to 82.3) and guideline adherence (68.5 to 74.5), with corresponding gains on all out-of-domain benchmarks. Increasing the flow length beyond four nodes provides little additional benefit: adherence remains comparable at $N=6$ (74.4), while weighted accuracy slightly decreases to 81.2. At $N=8$, both metrics decline further, with similar degradation on the out-of-domain benchmarks. These results suggest that moderately sized decision flows provide sufficient structure to capture clinically relevant decision steps, while overly short flows omit useful process information and longer flows introduce unnecessary complexity without consistent performance gains.

\begin{table*}[t]
  \centering
\caption{\textbf{Effect of the number of sampled decision flows on Qwen3-14B.}
We vary the number of controller-sampled flows $M$ used for Monte Carlo process supervision. We report weighted accuracy (WAcc.) and guideline adherence (Adh.) on MedGUIDE, aggregate score on AMEGA-LLM, and LLM-judge scores on RealMedQA and CancerGUIDE. Higher is better for all metrics.}
  \label{tab:main_results}

  \resizebox{\textwidth}{!}{%
  \begin{tabular}{lccccc}
  \toprule
  \multirow{2}{*}{\# Decision Flows (M)}
  & \multicolumn{2}{c}{MedGUIDE}
  & \multirow{2}{*}{AMEGA-LLM}
  & \multirow{2}{*}{RealMedQA}
  & \multirow{2}{*}{CancerGUIDE} \\
  \cmidrule(lr){2-3}
  & WAcc. & Adh.
  & & & \\

    \midrule
  1
  & 77.1 & 70.2 & 20.3 & 1.6 & 1.9 \\  
  2 
  & 80.3 & 72.5 & 21.6 & 1.7 & 2.1 \\
  4 
  & 82.5 & 73.5 & 22.3 & 1.9 & 2.4 \\
  6 
  & 82.4 & 74.1 & 22.1 & 1.9 & 2.3 \\
  8
  & 82.3 & 74.5 & 22.2 & 1.9 & 2.3 \\

  \bottomrule
  \end{tabular}%
  }
\end{table*}

\begin{table*}[t]
  \centering
\caption{\textbf{Effect of the number of decision nodes in sampled decision flows on  Qwen3-14B.} We vary the maximum number of decision nodes $N$ in each controller-sampled flow and evaluate its effect on in-domain and out-of-domain performance. We report weighted accuracy (WAcc.) and guideline adherence (Adh.) on MedGUIDE, aggregate score on AMEGA-LLM, and LLM-judge scores on RealMedQA and CancerGUIDE. Higher is better for all metrics.}
  \label{tab:main_results}

  \resizebox{\textwidth}{!}{%
  \begin{tabular}{lccccc}
  \toprule
  \multirow{2}{*}{\# Decision Nodes (N)}
  & \multicolumn{2}{c}{MedGUIDE}
  & \multirow{2}{*}{AMEGA-LLM}
  & \multirow{2}{*}{RealMedQA}
  & \multirow{2}{*}{CancerGUIDE} \\
  \cmidrule(lr){2-3}
  & WAcc. & Adh.
  & & & \\

    \midrule
    
  2 
  & 77.2 & 68.5 & 20.3 & 1.5 & 1.8 \\
  4 
  & 82.3 & 74.5 & 22.2 & 1.9 & 2.3 \\
  6 
  & 81.2 & 74.4 & 22.4 & 1.8 & 2.4 \\
  8
  & 80.5 & 72.8 & 21.5 & 1.8 & 2.1 \\

  \bottomrule
  \end{tabular}%
  }
\end{table*}

\subsection{Evaluation Templates}
\label{app:evaluation_prompts}

For open-ended and guideline-adherence evaluation, we follow the evaluation
templates and scoring rubrics provided by the respective benchmarks. We do
not modify the benchmark-specific evaluation criteria; we instantiate the
LLM judge with GPT-5.4. The templates used in our experiments are reproduced
below for completeness. Judges receive the benchmark-specified case,
reference evidence or criteria, and model output, but not the model identity
or training condition. The templates for MedGUIDE guideline-adherence, AMEGA-LLM, RealMedQA, and CancerGUIDE evaluation are provided in
Figs.~\ref{fig:template_medguide}, \ref{fig:template_amega},
\ref{fig:template_realmedqa}, and \ref{fig:template_cancerguide},
respectively.

\subsection{Controller and Policy Prompts}
We use separate prompts for the controller and policy, provided in
Figures~\ref{fig:inst_controller} and~\ref{fig:inst_policy}. Both roles receive
the same clinical question. The controller is prompted to generate plausible
guideline-style candidate decision flows, while the policy independently
generates its own decision flow followed by the final answer. The policy does
not observe the controller-generated flows; instead, they provide process-level
supervision through the reward. This separation provides process supervision
without directly constraining the policy to reproduce a particular trajectory.

\begin{figure}[t!]
\centering
\begin{controller}

\textbf{SYSTEM:}

You are a clinical decision-flow controller. Given a patient vignette, generate a compact decision flow that will answer the question.\\

\textbf{USER:}

Question: \{question\}\\

Generate the likely guideline-style decision flow. Use exactly this format:

Decision flow: \\
1. $<$short decision state$>$ \\
2. $<$short decision state$>$ \\

\end{controller}
\caption{\textbf{LLM instructions for the decision-flow controller.}
Given a clinical question, the controller generates a compact guideline-style decision flow. The generated flows represent plausible clinical decision pathways and are used to provide process-level supervision during policy optimization.}
\label{fig:inst_controller}
\end{figure}

\begin{figure}[t!]
\centering
\begin{policy}

\textbf{SYSTEM:}

You are taking a clinical multiple-choice benchmark. First generate a compact guideline-style decision flow, then give the final option letter.\\

\textbf{USER:}

Question: \{question\}

Your response must use exactly this format: \\

Decision flow: 

1. $<$short decision state$>$ 

2. $<$short decision state$>$ 

Answer: $<$single option letter$>$ \\

Use at most $N$ decision-flow nodes. Do not include guideline path text from memory; infer the flow from the vignette.

\end{policy}
\caption{\textbf{LLM instructions for the clinical reasoning policy.}
Given a clinical question, the policy generates its own compact guideline-style decision flow followed by the final answer. The policy does not observe controller-generated flows, allowing it to independently reason over the clinical case while receiving process-level supervision through the optimization objective. $N$ denotes the maximum number of decision-flow nodes.}
\label{fig:inst_policy}
\end{figure}

\begin{figure}[t!]
\centering
\begin{template}

Evaluate whether the model-generated clinical decision flow adheres to the
gold MedGUIDE guideline path for this case. \\

Judge semantic clinical equivalence and ordering, not exact wording. Do not
reward generic guideline-like statements unless they correspond to a gold
path step. Do not grade MCQ answer correctness as the main criterion; focus
on whether the decision flow uses the right clinical criteria and reaches a
path compatible with the gold trajectory. \\

Score rubric:

5 = all key gold decision states are captured in a clinically equivalent
    order, with no important contradictions.
    
4 = most key states are captured; only minor omissions or mild ambiguity.

3 = partially aligned but misses an important criterion, branch, or action.

2 = weak or mostly generic alignment; little specific path adherence.

1 = mostly wrong path or clinically contradicted by the gold path.

0 = empty or unusable decision flow. \\

Return \textbf{JSON}:  "score": ;
\end{template}
\caption{\textbf{MedGUIDE Guideline Adherence Judge Template.}
For each MedGUIDE example, the judge receives the patient vignette, the
supervised gold decision path, the gold answer option, the model's parsed
answer, and the model-generated decision flow.}
\label{fig:template_medguide}
\end{figure}

\subsection{Qualitative Analysis}
In the MedGUIDE example (Figure~\ref{fig:example_medguide}), the model identifies the key decision states---relapse after systemic therapy and poor performance status (ECOG PS 3)---and generates a concise decision flow consistent with the gold guideline recommendation, ultimately selecting the correct treatment option. This demonstrates that the model can recover clinically relevant decision states without requiring exact reproduction of the reference flow.

\begin{figure}[t!]
\centering
\begin{template}

Evaluate whether the model answer satisfies each AMEGA scoring criterion. \\

Rules:

- Judge only the model answer against the criterion and the case/question.

- Return true only when the answer clearly satisfies the criterion.

- Return false for missing, contradicted, or ambiguous information.

- Do not award credit for unsafe or clinically inappropriate statements.

Return exactly one Boolean verdict per criterion:
{"verdicts": [true, false, ...]}

Each true verdict receives the criterion's predefined weight. We report the aggregate score on a 0--50 scale:

\begin{equation}
    \mathrm{AMEGA}_{50}
    =
    50
    \frac{\sum_i s_i}{\sum_i s_i^{\max}}.
\end{equation}

Return \textbf{JSON}:  "score": ;
\end{template}
\caption{\textbf{AMEGA judge template.} AMEGA-LLM provides a set of weighted clinical criteria for each case--question pair. The judge evaluates each criterion independently.}
\label{fig:template_amega}
\end{figure}

\begin{figure}[t!]
\centering
\begin{template}

You evaluate whether a model-generated clinical decision/support flow is
supported by a NICE guideline recommendation. Use only the provided clinical
question, reference recommendation, and model output.

Score the model output against the reference NICE recommendation.

Scoring rubric:

5 = fully answers the question and is faithful to the recommendation.

4 = mostly faithful, with only minor omissions.

3 = partially supported or incomplete.

2 = weakly related, with important missing or unsupported components.

1 = mostly unsupported or misleading.

0 = contradicted, unsafe, or unrelated.

Also judge whether the decision flow explicitly connects the clinical
question to the recommendation. \\

Return \textbf{JSON}:  "score": ;
\end{template}
\caption{\textbf{RealMedQA judge template.} For RealMedQA, the judge compares the generated clinical decision flow with the reference NICE recommendation.}
\label{fig:template_realmedqa}
\end{figure}

\begin{figure}[t!]
\centering
\begin{template}

You evaluate NSCLC treatment recommendation predictions against a reference
CancerGUIDE recommendation. Use only the patient profile, reference
recommendation, and model output.

Score whether the model's final recommendation is clinically equivalent to
the reference recommendation.

Scoring rubric:

5 = fully answers the question and is faithful to the recommendation.

4 = mostly faithful, with only minor omissions.

3 = partially supported or incomplete.

2 = weakly related, with important missing or unsupported components.

1 = mostly unsupported or misleading.

0 = contradicted, unsafe, or unrelated.

Also judge whether the decision flow explicitly connects the clinical
question to the recommendation. \\

Return \textbf{JSON}:  "score": ;
\end{template}
\caption{\textbf{CancerGuide judge template.} For CancerGUIDE, the judge evaluates whether the generated cancer-treatment recommendation is clinically equivalent to the reference recommendation.}
\label{fig:template_cancerguide}
\end{figure}

\begin{figure}[t!]
\centering
\begin{example}

A 72-year-old woman with extensive-stage small-cell lung cancer initially responded to cisplatin/etoposide but relapsed after six months. Second-line topotecan also failed after four months. She now has progressive fatigue, weight loss, dyspnea, new liver and bone metastases, worsening liver function, and ECOG performance status 3.

What treatment option should be considered?

A) Palliative and best supportive care, with consideration of single-agent chemotherapy or palliative radiotherapy

B) Continue systemic therapy

C) Clinical trial

D) Continued surveillance

E) Palliative and best supportive care

F) Palliative/supportive care with possible chemotherapy, targeted therapy, or radiotherapy

G) Adjuvant therapy

\textbf{Correct answer:}

A — Palliative and best supportive care, with consideration of single-agent chemotherapy or palliative radiotherapy. \\ \\

\textbf{Gold guideline recommendation:}

Poor performance status. Palliative and best supportive care ± single-agent chemotherapy/palliative radiotherapy \\ \\

\textbf{Generated decision flow:}

1. Relapse or primary progressive disease after systemic therapy for ES-SCLC

2. Poor performance status (ECOG PS 3)

Final answer: A

\end{example}
\caption{\textbf{Example clinical question and generated decision flow from MedGUIDE.}
Given a patient with relapsed extensive-stage small-cell lung cancer and poor performance status, the model generates a compact decision flow identifying the clinically relevant guideline conditions and arrives at the correct treatment recommendation. The generated flow is consistent with the gold guideline recommendation while expressing it as a concise sequence of decision states.}
\label{fig:example_medguide}
\end{figure}

\end{document}